\documentclass[final]{cvpr}

\usepackage{times}
\usepackage{caption}
\usepackage{subcaption}
\usepackage{epsfig}
\usepackage{graphicx}
\usepackage{amsmath}
\usepackage{amssymb}
\usepackage{array}
\usepackage{colortbl}
\usepackage{booktabs}
\usepackage{bbm}
\usepackage{multirow}
\usepackage{multicol}
\usepackage{makecell}
\usepackage{xcolor}
\usepackage{xspace}
\usepackage{float}
\usepackage{color}

\usepackage{pifont}

\usepackage{epsfig}
\usepackage{graphicx}
\usepackage{amsmath}
\usepackage{amssymb}

\usepackage{booktabs}
\usepackage{tabulary}
\usepackage{dblfloatfix}
\usepackage{transparent}

\usepackage{algorithm}
\usepackage{listings}
\usepackage{algorithmic}
\usepackage{balance}

\usepackage[margin=4pt,font=small,labelfont=bf,labelsep=endash,tableposition=top]{caption}

\usepackage{etoolbox}
\makeatletter
\AfterEndEnvironment{algorithm}{\let\@algcomment\relax}
\AtEndEnvironment{algorithm}{\kern2pt\hrule\relax\vskip3pt\@algcomment}
\let\@algcomment\relax
\newcommand\algcomment[1]{\def\@algcomment{\footnotesize#1}}
\renewcommand\fs@ruled{\def\@fs@cfont{\bfseries}\let\@fs@capt\floatc@ruled
  \def\@fs@pre{\hrule height.8pt depth0pt \kern2pt}%
  \def\@fs@post{}%
  \def\@fs@mid{\kern2pt\hrule\kern2pt}%
  \let\@fs@iftopcapt\iftrue}
\makeatother

\definecolor{codegreen}{rgb}{0,0.5,0}
\definecolor{codeblue}{rgb}{0.25,0.5,0.5}
\definecolor{codegray}{rgb}{0.6,0.6,0.6}

\usepackage[pagebackref=true,breaklinks=true,colorlinks,bookmarks=false]{hyperref}

\def\cvprPaperID{0000}
\def\confYear{CVPR 2021}

\usepackage{bm}
\newcommand{\myparagraph}[1]{{\vspace{.5em} \noindent \bf #1}}
\newcommand{\app}{\raise.17ex\hbox{$\scriptstyle\sim$}}
\def\Ours{{Sparse R-CNN}\xspace}

\begin{document}

\title{Sparse R-CNN: End-to-End Object Detection with Learnable Proposals}

\author
{
Peize Sun$^{1*}$, 
~~~
Rufeng Zhang$^{2*}$, 
~~~
Yi Jiang$^{3*}$, 
~~~
Tao Kong$^{3}$, 
~~~
Chenfeng Xu$^{4}$,
~~~
Wei Zhan$^{4}$,\\
~~~
Masayoshi Tomizuka$^{4}$, 
~~~
Lei Li$^{3}$, 
~~~
Zehuan Yuan$^{3}$, 
~~~
Changhu Wang$^{3}$, 
~~~
Ping Luo$^{1}$
\\[0.2cm]
${^1}$The University of Hong Kong ~~~
${^2}$Tongji University\\
${^3}$ByteDance AI Lab ~~~
${^4}$University of California, Berkeley
}

\twocolumn[{
\maketitle
\vspace{-10mm}
\begin{figure}[H]
\hsize=\textwidth
\centering
\hspace{-0.2in}
\begin{subfigure}{0.30\textwidth}
    \centering
    \includegraphics[width=1.00\textwidth]{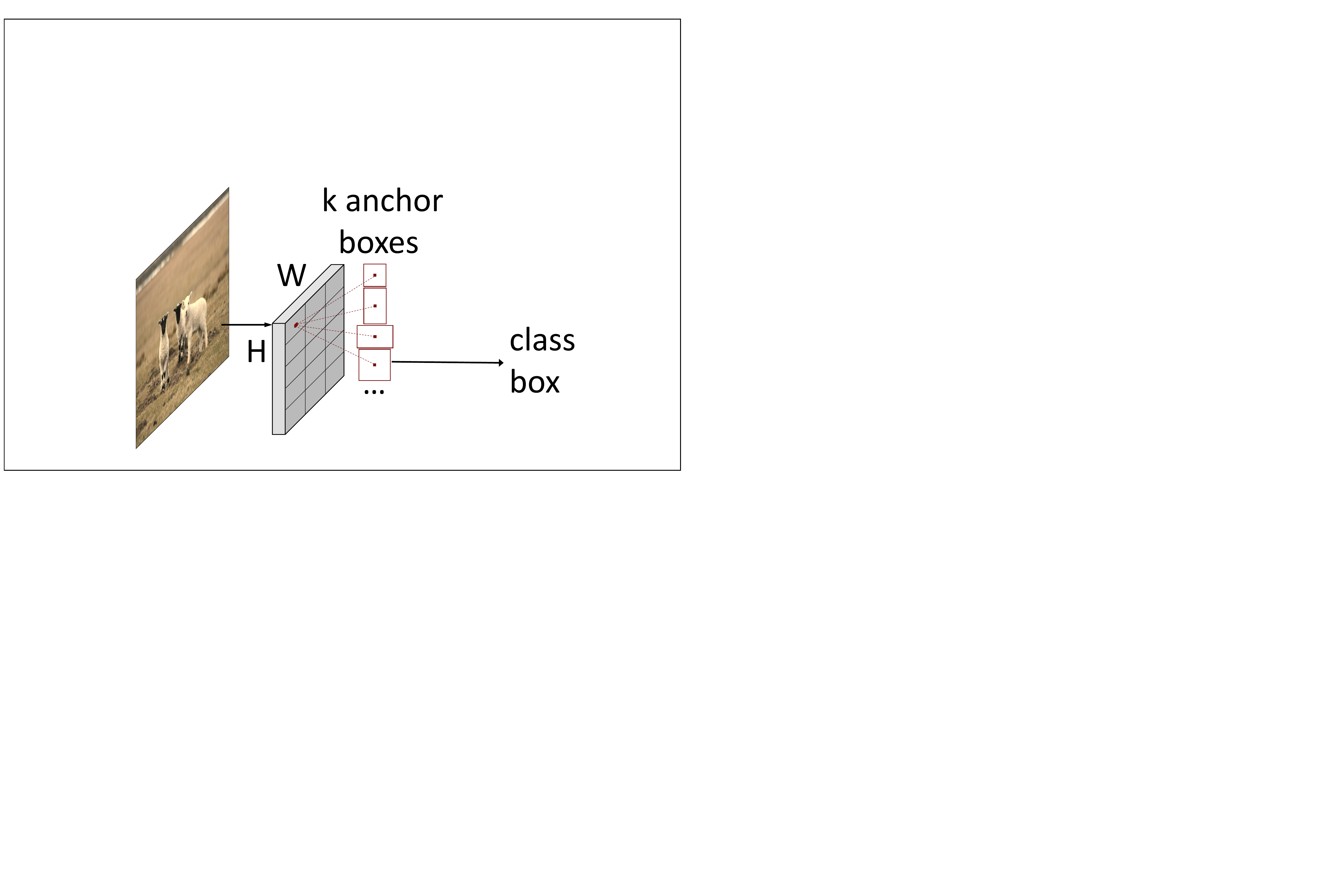}
    \caption{\textbf{Dense}: RetinaNet}
    \label{fig:1a}	
\end{subfigure}    
\begin{subfigure}{0.30\textwidth}
     \centering
     \includegraphics[width=1.00\textwidth]{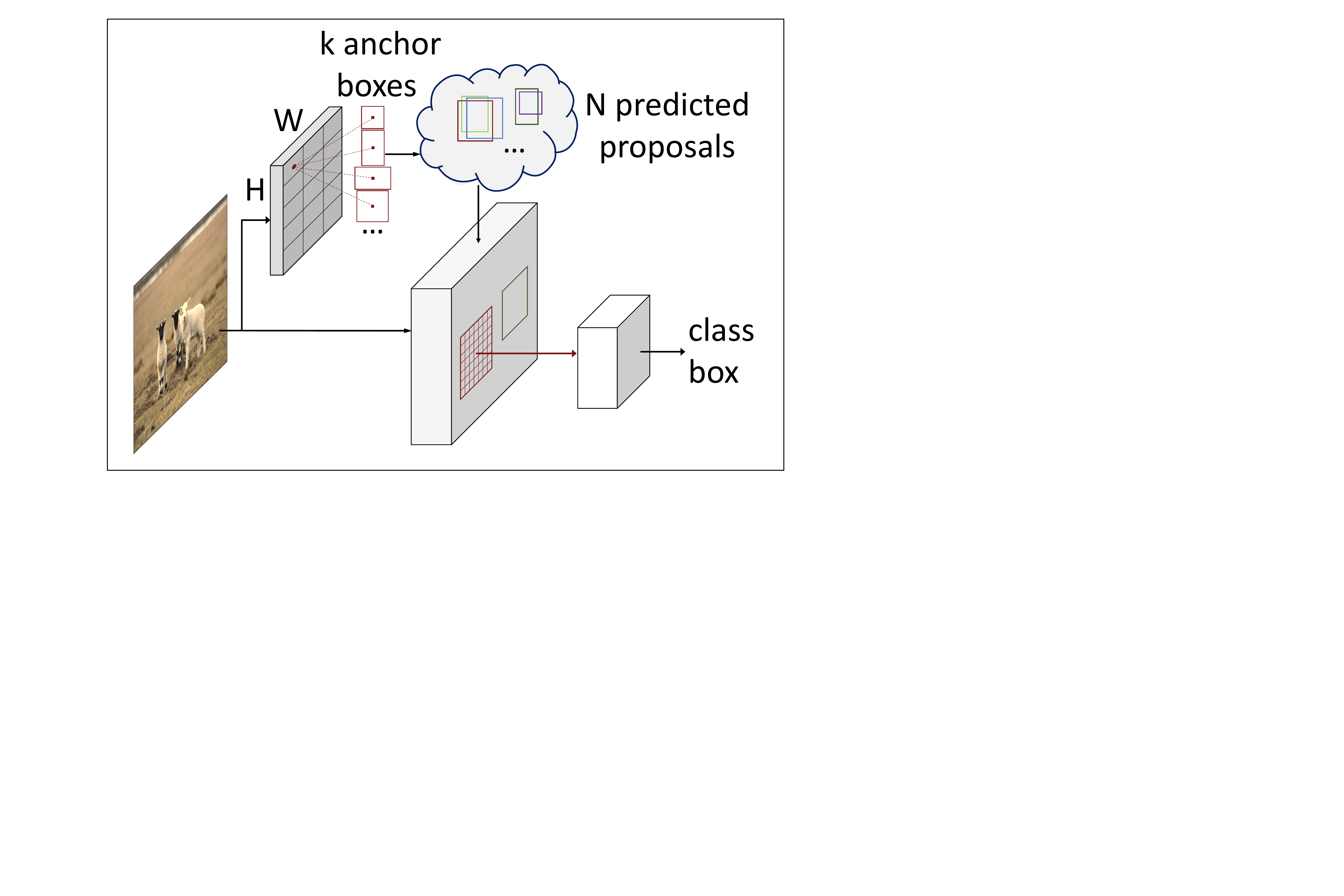}
     \caption{\textbf{Dense-to-Sparse}: Faster R-CNN}
     \label{fig:1b}
\end{subfigure}
\hspace{0.2in}
\begin{subfigure}{0.30\textwidth}
     \centering
     \includegraphics[width=1.00\textwidth]{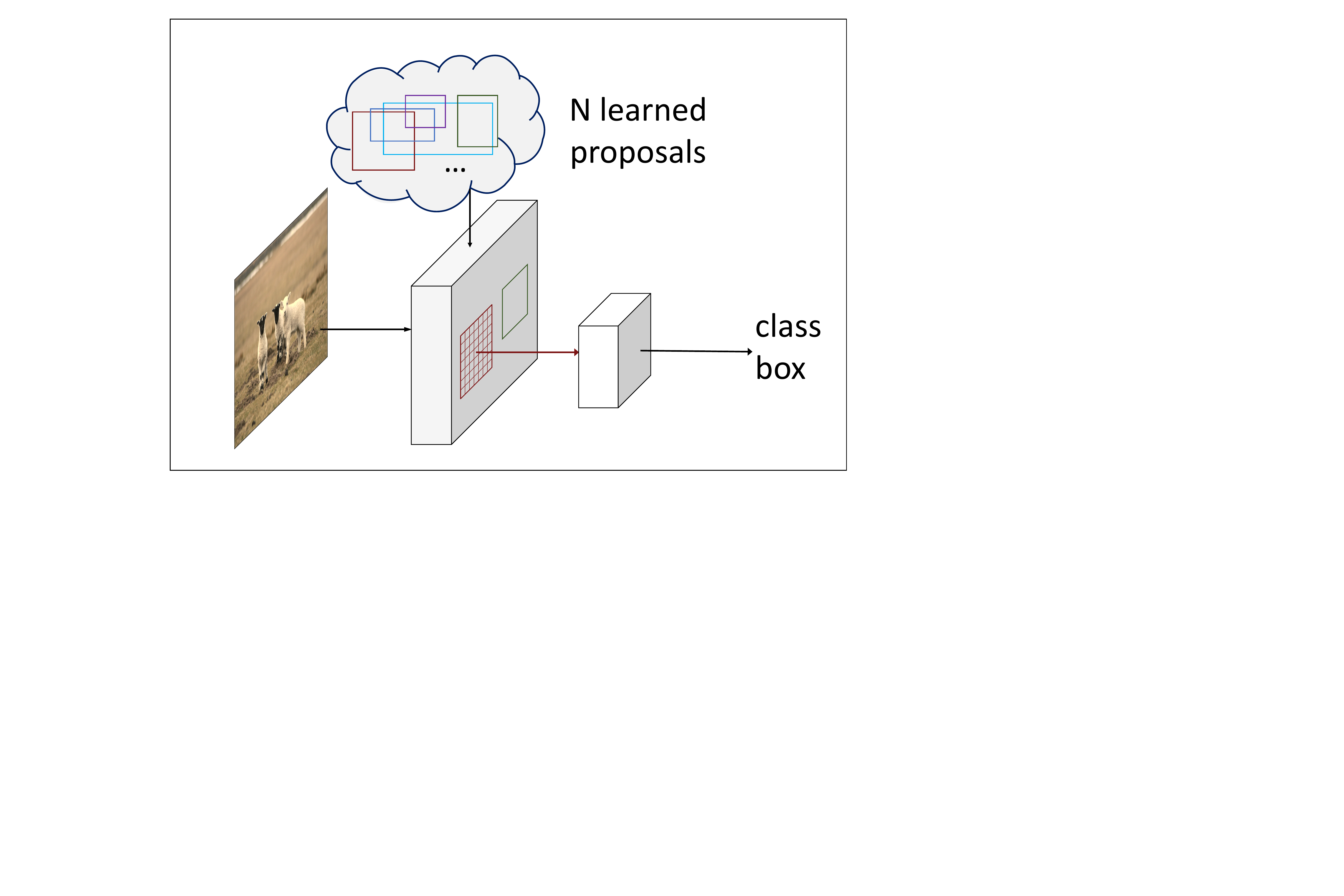}
     \caption{\textbf{Sparse}: Sparse R-CNN}
     \label{fig:1d}
\end{subfigure}
\caption{\textbf{Comparisons} of different object detection pipelines. 
(a) In dense detectors,  $HWk$ object candidates enumerate on all image grids, \eg RetinaNet~\cite{FocalLoss}.
(b) In dense-to-sparse detectors, they select a small set of $N$ candidates from dense $HWk$ object candidates, and then extract image features within corresponding regions by pooling operation, \eg Faster R-CNN~\cite{FasterRCNN}.
(c) Our proposed Sparse R-CNN,  directly provides a small set of $N$ learned object proposals.
Here \textbf{$N \ll HWk$}.
}
\label{fig:1}
\end{figure}
\vspace{4mm}
}]

\footnotetext{* Equal contribution.}
\pagestyle{empty}
\thispagestyle{empty}

\begin{abstract}
\vspace{-2mm}
We present Sparse R-CNN, a purely sparse method for object detection in images. Existing works on object detection heavily rely on dense object candidates, such as $k$ anchor boxes pre-defined on all grids of image feature map of size $H\times W$. In our method, however, a fixed sparse set of learned object proposals, total length of $N$, are provided to object recognition head to perform classification and location. By eliminating $HWk$ (up to hundreds of thousands) hand-designed object candidates to $N$ (\eg 100) learnable proposals, Sparse R-CNN completely avoids all efforts related to object candidates design and many-to-one label assignment. More importantly, final predictions are directly output without non-maximum suppression post-procedure. Sparse R-CNN demonstrates accuracy, run-time and training convergence performance on par with the well-established  detector baselines on the challenging COCO dataset, \eg, achieving 45.0 AP in standard $3\times$ training schedule and running at 22 fps using ResNet-50 FPN model. We hope our work could inspire re-thinking the convention of dense prior in object detectors. The code is available at: {\footnotesize{\url{https://github.com/PeizeSun/SparseR-CNN}}}.
\end{abstract}

\begin{figure}[t]
\vspace{-3mm}
\hspace{-2mm}
\includegraphics[width=0.55\textwidth]{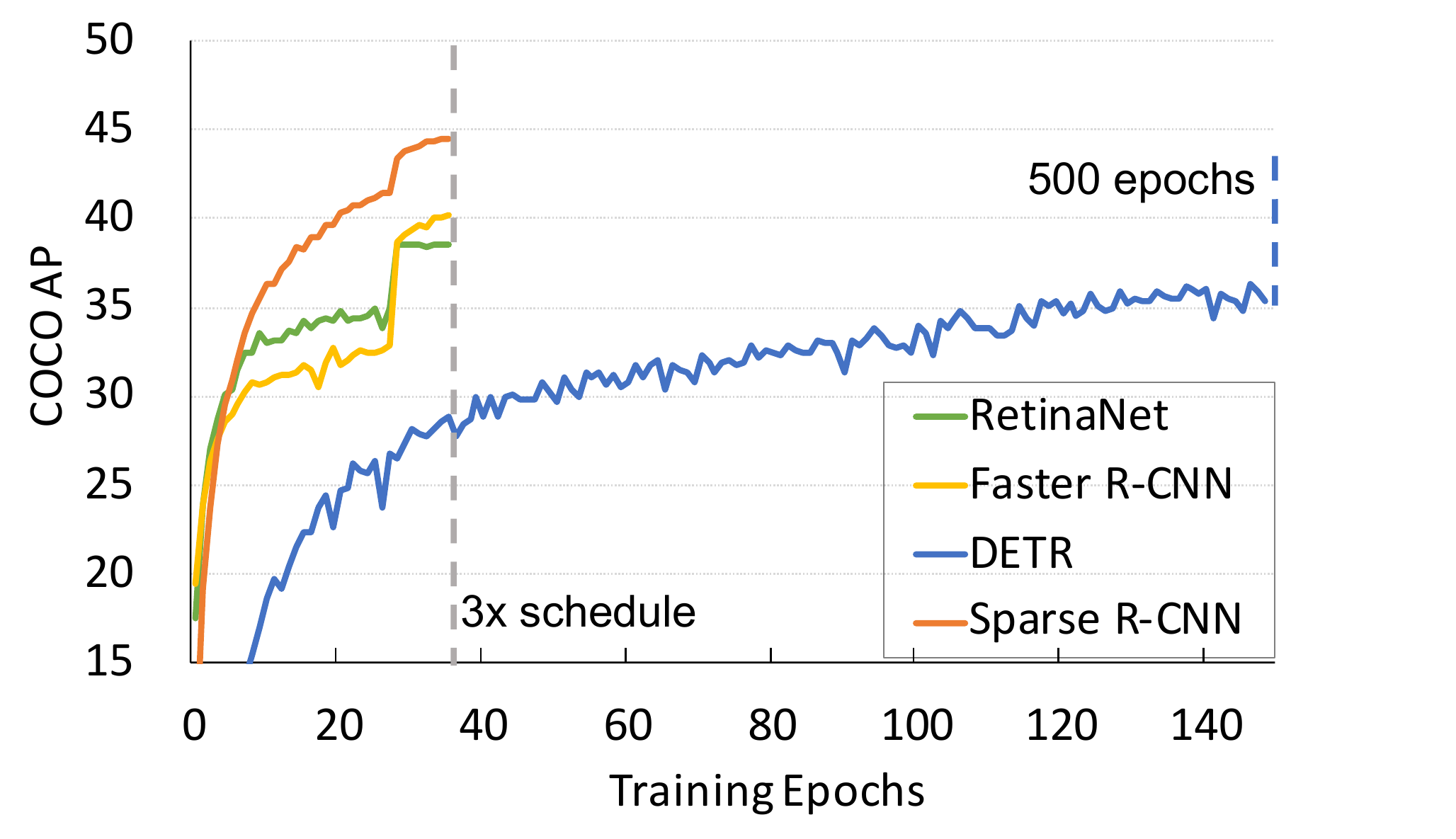}
\caption{Convergence curves of RetinaNet, Faster R-CNN, DETR and Sparse R-CNN on COCO \texttt{val2017}~\cite{COCO}. Sparse R-CNN achieves competitive performance in terms of training efficiency and detection quality.}
\label{fig:curve}
\vspace{-2mm}
\end{figure}

\section{Introduction}

Object detection aims at localizing a set of objects and recognizing their categories in an image. 
Dense prior has always been cornerstone to success in detectors. 
In classic computer vision, the sliding-window paradigm, in which a classifier is applied on a dense image grid, is leading detection method for decades~\cite{hog, dpm, boosted}. 
Modern mainstream one-stage detectors pre-define marks on a dense feature map grid, such as anchors boxes~\cite{FocalLoss, YOLO9000}, shown in Figure~\ref{fig:1a}, or reference points~\cite{FCOS, CenterNet}, and predict the relative scaling and offsets to bounding boxes of objects, as well as the corresponding categories. Although two-stage pipelines work on a sparse set of proposal boxes, their proposal generation algorithms are still built on dense candidates~\cite{RCNN, FasterRCNN}, shown in Figure~\ref{fig:1b}.

These well-established methods are conceptually intuitive and offer robust performance~\cite{PASCAL-VOC, COCO}, together with fast training and inference time~\cite{detectron2}. Besides their great success, it is important to note that dense-prior detectors suffer some limitations:
1) Such pipelines usually produce redundant and near-duplicate results, thus making non-maximum suppression (NMS)~\cite{Soft-NMS, SOLOv2} post-processing a necessary component. 
2) The many-to-one label assignment problem~\cite{CascadeRCNN, DynamicRCNN, ATSS} in training makes the network sensitive to heuristic assign rules. 
3) The final performance is largely affected by sizes, aspect ratios and number of anchor boxes~\cite{FocalLoss, YOLO9000}, density of reference points~\cite{foveabox, FCOS, CenterNet} and proposal generation algorithm~\cite{RCNN, FasterRCNN}.

Despite the dense convention is widely recognized among object detectors, a natural question to ask is: \textit{Is it possible to design a sparse detector?} Recently, DETR proposes to reformulate object detection as a direct and sparse set prediction problem~\cite{DETR}, whose input is merely 100 learned object queries~\cite{vaswani2017attention}.
The final set of predictions are output directly without any hand-designed post-processing. In spite of its simple and fantastic framework, DETR requires each object query to interact with global image context. This dense property not only slows down its training convergence~\cite{deformabledetr}, but also blocks it establishing a thoroughly sparse pipeline for object detection.

We believe the sparse property should be in two aspects: \textit{sparse boxes} and \textit{sparse features}. 
Sparse boxes mean that a small number of starting boxes (\eg 100) is enough to predict all objects in an image.
While sparse features indicate the feature of each box does not need to interact with all other features over the full image.
From this perspective, 
DETR is not a pure sparse method since each object query must interact with dense features over full images.

In this paper, we propose Sparse R-CNN, a purely sparse method, without object positional candidates enumerating on \textit{all(dense) image grids} nor object queries interacting with \textit{global(dense) image feature}. 
As shown in Figure~\ref{fig:1d}, object candidates are given with a fixed small set of learnable bounding boxes represented by 4-d coordinate. For example of COCO dataset~\cite{COCO}, 100 boxes and 400 parameters are needed in total, rather than the predicted ones from hundreds of thousands of candidates in Region Proposal Network (RPN)~\cite{FasterRCNN}. 
These sparse candidates are used as proposal boxes to extract the feature of Region of Interest (RoI) by RoIPool~\cite{FastRCNN} or RoIAlign~\cite{MaskRCNN}.

The learnable proposal boxes are the statistics of potential object location in the image. Whereas, the 4-d coordinate is merely a rough representation of object and lacks a lot of informative details such as pose and shape.
Here we introduce another key concept termed \textit{proposal feature}, which is a high-dimension (\eg, 256) latent vector. Compared with rough bounding box, it is expected to encode the rich instance characteristics. Specially, proposal feature generates a series of customized parameters for its exclusive object recognition head. We call this operation Dynamic Instance Interactive Head, since it shares similarities with recent dynamic scheme~\cite{jia2016dynamic, condinst}.
Compared to the shared 2-fc layers in ~\cite{FasterRCNN}, our head is more flexible and holds a significant lead in accuracy. We show in our experiment that the formulation of head conditioned on unique proposal feature instead of the fixed parameters is actually the key to Sparse R-CNN's success.
Both \textit{proposal boxes} and \textit{proposal features} are randomly initialized and optimized together with other parameters in the whole network.

The most remarkable property in our Sparse R-CNN is its sparse-in sparse-out paradigm in the whole time. The initial input is a sparse set of proposal boxes and proposal features, together with the one-to-one dynamic instance interaction. Neither dense candidates~\cite{FocalLoss, FasterRCNN} nor interacting with global(dense) feature~\cite{DETR} exists in the pipeline. 
This pure sparsity makes Sparse R-CNN a brand new member in R-CNN family. 

Sparse  R-CNN  demonstrates its accuracy, run-time and  training convergence performance on par with the well-established detectors~\cite{CascadeRCNN, FasterRCNN, FCOS} on the challenging COCO dataset~\cite{COCO}, \eg, achieving 45.0 AP in standard $3\times$ training schedule and running at 22 fps using ResNet-50 FPN model.
To our best knowledge, the proposed Sparse R-CNN is the first work that demonstrates a considerably sparse design is qualified yet. We hope our work could inspire re-thinking the necessary of dense prior in object detection and exploring next generation of object detector.

\section{Related Work}
\myparagraph{Dense method.} Sliding-window paradigm has been popular for many years in object detection. 
Limited by classical feature extraction techniques~\cite{hog, boosted}, the performance has plateaued for decades and the application scenarios are limited.  
Development of deep convolution neural networks (CNNs)~\cite{ResNet, BatchNorm, AlexNet} cultivates general object detection achieving significant improvement in performance~\cite{PASCAL-VOC, COCO}. 
One of mainstream pipelines is one-stage detector, which directly predicts the category and location of anchor boxes densely covering spatial positions, scales, and aspect ratios in a single-shot way, such as OverFeat~\cite{OverFeat}, YOLO~\cite{YOLO9000}, SSD~\cite{SSD} and RetinaNet~\cite{FocalLoss}. Recently, anchor-free algorithms~\cite{DenseBox, CornerNet, FCOS, CenterNet, foveabox} are proposed to make this pipeline much simpler by replacing hand-crafted anchor boxes with reference points. All of above methods are built on dense candidates and each candidate is directly classified and regressed. These candidates are assigned to ground-truth object boxes in training time based on a pre-defined principle, \eg, whether the anchor has a higher intersection-over-union (IoU) threshold with its corresponding ground truth, or whether the reference point falls in one of object boxes. Moreover, NMS post-processing~\cite{Soft-NMS, SOLOv2} is needed to remove redundant predictions during inference time.

\myparagraph{Dense-to-sparse method.} Two-stage detector is another mainstream pipeline and has dominated modern object detection for years~\cite{CascadeRCNN, R-FCN, FastRCNN, RCNN, FasterRCNN}. 
This paradigm can be viewed as an extension of dense detector. It first obtains a sparse set of foreground proposal boxes from dense region candidates, and then refines location of each proposal and predicts its specific category. The region proposal algorithm plays an important role in the first stage in these two-stage methods, such as Selective Search~\cite{SelectiveSearch} in R-CNN and Region Proposal Networks (RPN)~\cite{FasterRCNN} in Faster R-CNN. Similar to dense pipeline, it also needs NMS post-processing and hand-crafted label assignment. There are only a few of foreground proposals from hundreds of thousands of candidates, thus these detectors can be concluded as dense-to-sparse methods.

Recently, DETR~\cite{DETR} is proposed to directly output the predictions without any hand-crafted components, achieving promising performance. DETR utilizes a sparse set of object queries, to interact with global(dense) image feature, in this view, it can be seen as another dense-to-sparse formulation.

\myparagraph{Sparse method.} Sparse object detection has the potential to eliminate efforts to design dense candidates, but usually has trailed the accuracy of above dense detectors.
G-CNN~\cite{gcnn} can be viewed as a precursor to this group of algorithms. It starts with a multi-scale regular grid over the image and iteratively updates the boxes to cover and classify objects. This hand-designed regular prior is obviously sub-optimal and fails to achieve top performance. Instead, our Sparse R-CNN applies learnable proposals and achieves better performance. Concurrently, Deformable-DETR~\cite{deformabledetr} is introduced to restrict  each object query to attend to a small set of key sampling points around the reference points, instead of all points in feature map. 
We hope sparse methods could serve as solid baseline and help ease future research in object detection community.

\section{\Ours}
\label{sec:3}
\begin{figure}[!t]
\begin{center}
\includegraphics[width=0.45\textwidth]{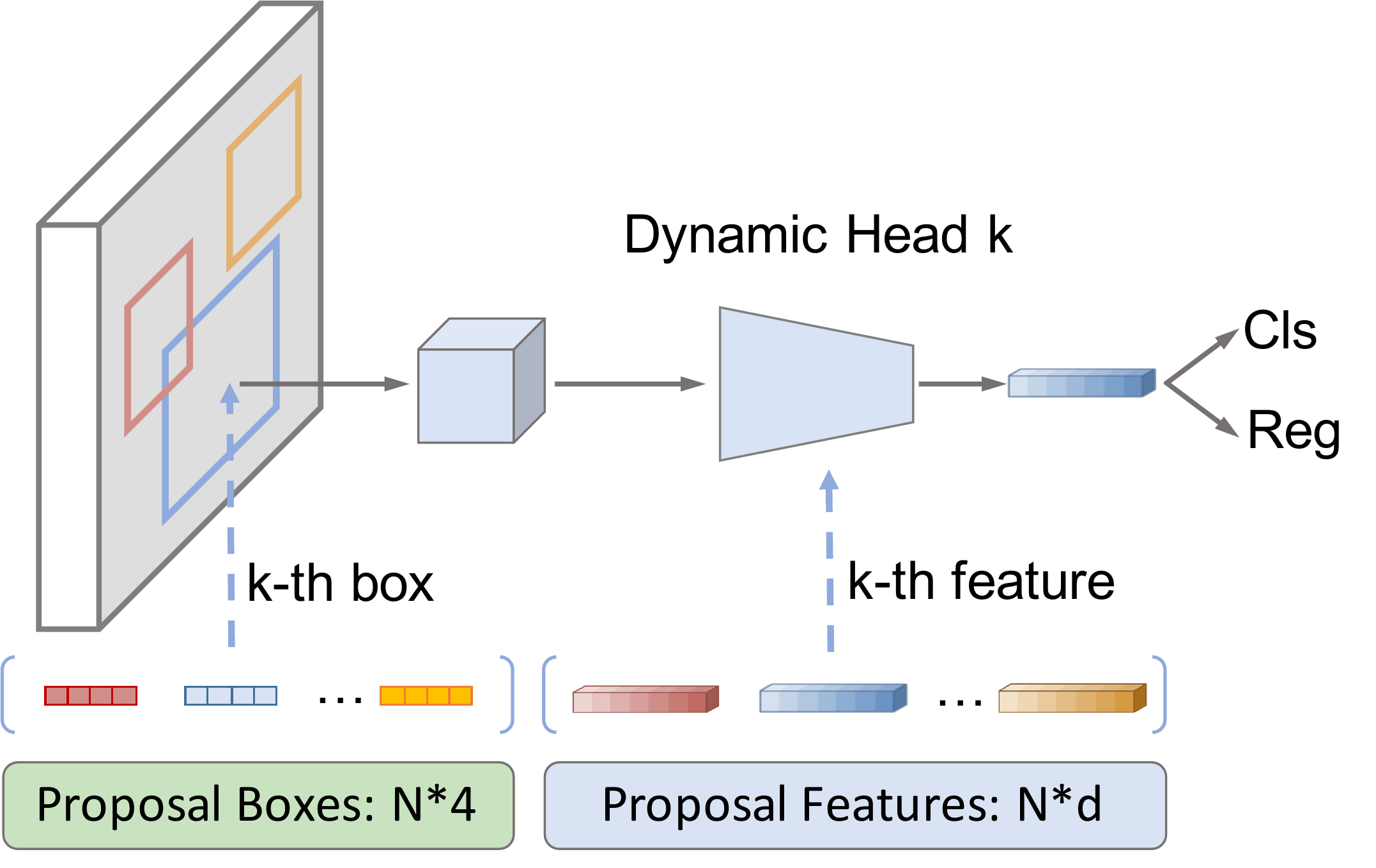}
\caption{An overview of Sparse R-CNN pipeline. The input includes an image, a set of proposal boxes and proposal features, where the latter two are learnable parameters. The backbone extracts feature map, each proposal box and proposal feature are fed into its exclusive dynamic head to generate object feature, and finally outputs classification and location.} 
\label{fig:pipeline}
\end{center}
\end{figure}

The key idea of Sparse R-CNN framework is to replace hundreds of thousands of candidates from Region Proposal Network (RPN) with a small set of proposal boxes (\eg, 100). The pipeline is shown in Figure~\ref{fig:pipeline}.

Sparse R-CNN is a simple, unified network composed of a backbone network, a dynamic instance interactive head and two task-specific prediction layers.
There are three inputs in total, an image, a set of proposal boxes and proposal features. The latter two are learnable and can be optimized together with other parameters in network. 
We will describe each components in this section in details.

\myparagraph{Backbone.}
Feature Pyramid Network (FPN) based on ResNet architecture~\cite{ResNet, FPN} is adopted as the backbone network to produce multi-scale feature maps from input image.
Following ~\cite{FPN}, we construct the pyramid with levels $P_2$ through $P_5$, where $l$ indicates pyramid level and $P_l$ has resolution $2^l$ lower than the input. All pyramid levels have $C = 256$ channels. Please refer to ~\cite{FPN} for more details. Actually, Sparse R-CNN has the potential to benefit from more complex designs to further improve its performance, such as stacked encoder layers~\cite{DETR} and deformable convolution network~\cite{DCN}, on which a recent work Deformable-DETR~\cite{deformabledetr} is built. However, we align the setting with Faster R-CNN~\cite{FasterRCNN, FPN} to show the simplicity and effectiveness of our method.

\myparagraph{Learnable proposal box.}
A fixed small set of learnable proposal boxes ($N \times 4$) are used as region proposals, instead of the predictions from Region Proposal Network (RPN).
These proposal boxes are represented by 4-d parameters ranging from 0 to 1, denoting normalized center coordinates, height and width.
The parameters of proposal boxes will be updated with the back-propagation algorithm during training.
Thanks to the learnable property, we find in our experiment that the effect of initialization is minimal, thus making the framework much more flexible.

Conceptually, these learned proposal boxes are the statistics of potential object location in the training set and can be seen as an initial guess of the regions that are most likely to encompass the objects in the image, regardless of the input.
Whereas, the proposals from RPN are strongly correlated to the current image and provide coarse object locations. We rethink that the first-stage locating is luxurious in the presence of later stages to refine the location of boxes. Instead, a reasonable statistic can already be qualified candidates. 
In this view, Sparse R-CNN can be categorized as the extension of object detector paradigm from thoroughly dense~\cite{FocalLoss, SSD, YOLO, FCOS} to dense-to-sparse~\cite{CascadeRCNN, R-FCN, RCNN, FasterRCNN} to thoroughly sparse, shown in Figure~\ref{fig:1}.

\myparagraph{Learnable proposal feature.}
Though the 4-d proposal box is a brief and explicit expression to describe objects, it provides a coarse localization of objects and a lot of informative details are lost, such as object pose and shape. 
Here we introduce another concept termed \textit{proposal feature} ($N \times d$), it is a high-dimension (\eg, 256) latent vector and is expected to encode the rich instance characteristics. The number of proposal features is same as boxes, and we will discuss how to use it next.

\lstset{
  backgroundcolor=\color{white},
  basicstyle=\fontsize{7.5pt}{8.5pt}\fontfamily{lmtt}\selectfont,
  columns=fullflexible,
  breaklines=true,
  captionpos=b,
  commentstyle=\fontsize{8pt}{9pt}\color{codegray},
  keywordstyle=\fontsize{8pt}{9pt}\color{codegreen},
  stringstyle=\fontsize{8pt}{9pt}\color{codeblue},
  frame=tb,
  otherkeywords = {self},
}
\begin{figure}[t]
\begin{lstlisting}[language=python]
def dynamic_instance_interaction(pro_feats, roi_feats):
    # pro_feats: (N, C)
    # roi_feats: (N, S*S, C)

    # parameters of two 1x1 convs: (N, 2 * C * C/4)
    dynamic_params = linear1(pro_features)
    # parameters of first conv: (N, C, C/4)
    param1 = dynamic_params[:, :C*C/4].view(N, C, C/4)
    # parameters of second conv: (N, C/4, C)
    param2 = dynamic_params[:, C*C/4:].view(N, C/4, C)
    
    # instance interaction for roi_features: (N, S*S, C)
    roi_feats = relu(norm(bmm(roi_feats, param1)))    
    roi_feats = relu(norm(bmm(roi_feats, param2)))
    
    # roi_feats are flattened: (N, S*S*C)
    roi_feats = roi_feats.flatten(1)
    # obj_feats: (N, C)
    obj_feats = linear2(roi_feats)
    return obj_feats
    
\end{lstlisting}
\vspace{-1em}
\caption{Pseudo-code of dynamic instance interaction, the $k$-th proposal feature generates dynamic parameters for the corresponding $k$-th RoI. \texttt{bmm}: batch matrix multiplication; \texttt{linear}: linear projection. }
\label{fig:head}
\vspace{3mm}
\end{figure}

\myparagraph{Dynamic instance interactive head.} 

Given $N$ proposal boxes, \Ours first utilizes the RoIAlign operation to extract features for each box. Then each box feature will be used to generate the final predictions using our prediction head. Motivated by dynamic algorithms~\cite{jia2016dynamic, condinst}, we propose Dynamic Instance Interactive Head. Each RoI feature is fed into its own exclusive head for object location and classification, where each head is conditioned on specific proposal feature.

Figure~\ref{fig:head} illustrates the  dynamic instance interaction. In our design, proposal feature and proposal box are in one-to-one correspondence.  For $N$ proposal boxs, $N$ proposal features are employed. Each RoI feature $f_i$($S\times S, C$) will interact with the corresponding proposal feature $p_i$($C$) to filter out ineffective bins and outputs the final object feature ($C$). For light design, we carry out two consecutive $1 \times 1$ convolutions with ReLU activation function, to implement the interaction process. The parameters of these two convolutions are generated by 
corresponding proposal feature.

The implementation details of interactive head is not crucial as long as parallel operation is supported for efficiency. The final regression prediction is computed by a 3-layer perception, and classification prediction is by a linear projection layer.

We also adopt the iteration structure~\cite{CascadeRCNN} and self-attention module~\cite{vaswani2017attention} to further improve the performance. 
For iteration structure, the newly generated object boxes and object features will serve as the proposal boxes and proposal features of the next stage in iterative process. Thanks to the sparse property and light dynamic head, it introduces only a marginal computation overhead. 
Before dynamic instance interaction, self-attention module is applied to the set of object features to reason about the relations between objects. We note that~\cite{RelationNetworks} also utilizes self-attention module. However, it demands geometry attributes and complex rank feature in addition to object feature. Our module is much more simple and only takes object feature as input.

\begin{table*}[t]
\begin{center}
{\begin{tabular}{l |c c |c c c c c c| c}

\arrayrulecolor{white}\hline
\Xhline{2\arrayrulewidth}
\arrayrulecolor{white}\hline

Method &Feature& Epochs & AP & AP$_{50}$ & AP$_{75}$ & AP$_s$ & AP$_m$ & AP$_l$ & 
FPS\\
\arrayrulecolor{white}\hline
\arrayrulecolor{black}\hline
\arrayrulecolor{white}\hline
RetinaNet-R50~\cite{detectron2}  & FPN & 36  & 38.7 & 58.0 & 41.5 & 23.3 & 42.3 & 50.3 &   24\\
RetinaNet-R101~\cite{detectron2}  & FPN & 36  & 40.4 & 60.2 & 43.2 & 24.0 & 44.3 & 52.2 &   18\\
Faster R-CNN-R50~\cite{detectron2}  & FPN & 36  & 40.2 & 61.0 & 43.8 & 24.2 & 43.5 & 52.0 &  26\\
Faster R-CNN-R101~\cite{detectron2}  & FPN & 36  & 42.0 & 62.5 & 45.9 & 25.2 & 45.6 & 54.6 &  20\\
Cascade R-CNN-R50~\cite{detectron2}& FPN & 36 & 44.3 &62.2 &48.0&26.6&47.7&57.7& 19\\
DETR-R50~\cite{DETR}  & Encoder & 500 & 42.0 & 62.4 & 44.2 & 20.5 & 45.8 & 61.1 & 28 \\
DETR-R101~\cite{DETR}  & Encoder & 500 & 43.5 & 63.8 & 46.4 & 21.9 & 48.0 & 61.8 & 20 \\
DETR-DC5-R50~\cite{DETR}  & Encoder & 500 & 43.3 & 63.1 & 45.9 & 22.5 & 47.3 & 61.1 & 12 \\
DETR-DC5-R101~\cite{DETR}  & Encoder & 500 & 44.9 & {\bf 64.7} &47.7 &23.7 & {\bf 49.5} & {\bf 62.3} & 10 \\
Deformable DETR-R50~\cite{deformabledetr}  & DeformEncoder & 50 & 43.8 & 62.6 & 47.7 & 26.4 & 47.1 & 58.0 & 19 \\
\arrayrulecolor{white}\hline
\arrayrulecolor{black}\hline
\arrayrulecolor{white}\hline
Sparse R-CNN-R50  &FPN & 36  & 42.8 & 61.2 & 45.7 & 26.7 & 44.6 & 57.6 & 23 \\
Sparse R-CNN-R101  &FPN & 36 & 44.1 & 62.1 & 47.2 & 26.1 & 46.3 & 59.7& 19 \\
Sparse R-CNN*-R50 &FPN & 36&  45.0 & 63.4 & 48.2 &26.9 & 47.2 & 59.5 & 22 \\
Sparse R-CNN*-R101 &FPN &36 & {\bf 46.4} & 64.6 & {\bf 49.5} & {\bf 28.3} & 48.3 & 61.6 & 18\\
\arrayrulecolor{white}\hline
\Xhline{2\arrayrulewidth}
\arrayrulecolor{white}\hline

\end{tabular}}
\end{center}
\vspace{-2mm}
\caption{Comparisons with different object detectors on COCO 2017 val set. The top section shows results from Detectron2~\cite{detectron2} or original papers~\cite{DETR, deformabledetr}. 
Here ``$*$" indicates that the model is with 300 learnable proposal boxes and random crop training augmentation, similar to Deformable DETR~\cite{deformabledetr}. 
Run time is evaluated on NVIDIA Tesla V100 GPU.
}
\label{table5_1}
\vspace{-4mm}
\end{table*}

\begin{table*}[t]
\begin{center}
{\setlength{\tabcolsep}{3.0mm}

\begin{tabular}{l |c c |c c c c c c}

\arrayrulecolor{white}\hline
\Xhline{2\arrayrulewidth}
\arrayrulecolor{white}\hline

Method &Backbone& TTA  & AP & AP$_{50}$ & AP$_{75}$ & AP$_s$ & AP$_m$ & AP$_l$ \\
\arrayrulecolor{white}\hline
\arrayrulecolor{black}\hline
\arrayrulecolor{white}\hline
CornerNet~\cite{CornerNet} &  Hourglass-104 & &40.6 & 56.4 & 43.2 & 19.1& 42.8 & 54.3\\
CenterNet~\cite{CenterNet} & Hourglass-104 & & 42.1& 61.1 &45.9 &24.1 & 45.5 &52.8\\
RepPoint~\cite{reppoints} & ResNet-101-DCN & & 45.0 & 66.1 & 49.0 & 26.6 & 48.6 & 57.5\\
FCOS~\cite{FCOS} & ResNeXt-101-DCN & & 46.6 & 65.9 & 50.8 & 28.6 &49.1 & 58.6\\ 
ATSS~\cite{ATSS} & ResNeXt-101-DCN & \checkmark & 50.7 & 68.9 & 56.3 & 33.2 & 52.9 & 62.4\\
YOLO~\cite{yolov4scale} & CSPDarkNet-53 & & 47.5& 66.2& 51.7& 28.2& 51.2  & 59.8\\ 
EfficientDet~\cite{EfficientDet} & EfficientNet-B5 & & 51.5 &  70.5 & 56.1 & - & - & -\\ 
\arrayrulecolor{white}\hline
\arrayrulecolor{black}\hline
\arrayrulecolor{white}\hline
Sparse R-CNN  &ResNeXt-101 &   & 46.9 & 66.3 & 51.2 & 28.6 & 49.2 & 58.7 \\
Sparse R-CNN  &ResNeXt-101-DCN &   & 48.9 & 68.3 & 53.4& 29.9 & 50.9 & 62.4 \\
Sparse R-CNN  &ResNeXt-101-DCN & \checkmark  & 51.5 & 71.1 & 57.1 & 34.2 & 53.4 & 64.1 \\
\arrayrulecolor{white}\hline
\Xhline{2\arrayrulewidth}
\arrayrulecolor{white}\hline

\end{tabular}}
\end{center}
\vspace{-2mm}
\caption{Comparisons with different object detectors on COCO 2017 test-dev set. The top section shows results from original papers. ``TTA" indicates test-time augmentations, following the settings in~\cite{ATSS}.
}
\label{table_test}
\vspace{-4mm}
\end{table*}

\myparagraph{Set prediction loss.}
Sparse R-CNN applies set prediction loss~\cite{DETR, stewart2016end, yang2019learning3d} on the fixed-size set of predictions of classification and box coordinates. Set-based loss produces an optimal bipartite matching between predictions and ground truth objects. 
The matching cost is defined as follows:
\begin{equation}
    \label{total_loss}
    \mathcal{L} = \lambda_{cls}  \cdot  \mathcal{L}_{\mathit{cls}} + \lambda_{L1} \cdot \mathcal{L}_{\mathit{L1}} +
    \lambda_{giou} \cdot \mathcal{L}_{\mathit{giou}}
\end{equation}
Here 
$\mathcal{L}_{\mathit{cls}}$ is focal loss~\cite{FocalLoss} of predicted classifications and ground truth category labels, $\mathcal{L}_{\mathit{L1}}$ and $\mathcal{L}_{\mathit{giou}}$ are L1 loss and generalized IoU loss~\cite{GIoU} between normalized center coordinates and height and width of predicted boxes and ground truth box, respectively. 
$\lambda_{cls}$, $\lambda_{L1}$ and $\lambda_{giou}$ are coefficients of each component.
The training loss is the same as the matching cost except that only performed on matched pairs. The final loss is the sum of all pairs normalized by the number of objects inside the training batch.

R-CNN families~\cite{CascadeRCNN, ATSS} have always been puzzled by label assignment problem since many-to-one matching remains. Here we provide new possibilities that directly bypassing many-to-one matching and introducing one-to-one matching with set-based loss. This is an attempt towards exploring end-to-end object detection.

\section{Experiments}

\myparagraph{Dataset.}
Our experiments are conducted on the challenging MS COCO  benchmark~\cite{COCO} using the standard metrics for object detection.
All models are trained on the COCO \texttt{train2017} split ($\sim$118k images) and evaluated with \texttt{val2017} (5k images).

\begin{table*}[t]
\begin{center}
{\begin{tabular}
{
c c c l l l l l l 
}
\toprule
Sparse & Iterative & Dynamic & 
\multicolumn{1}{c}{AP} & 
\multicolumn{1}{c}{AP$_{50}$}  & 
\multicolumn{1}{c}{AP$_{75}$}  & 
\multicolumn{1}{c}{AP$_s$}  & 
\multicolumn{1}{c}{AP$_m$}  & 
\multicolumn{1}{c}{AP$_l$}  \\
\midrule
\checkmark &  &    &  18.5 & 35.0 & 17.7& 8.3 & 21.7 & 26.4\\
\checkmark &\checkmark &    & 
32.2  \textcolor{blue}{(+\textbf{13.7})}& 
47.5  \textcolor{gray}{(+\textbf{12.5})}& 
34.4 \textcolor{gray}{(+\textbf{16.7})}& 
18.2  \textcolor{gray}{(+\textbf{9.9})}& 
35.2  \textcolor{gray}{(+\textbf{13.5})}& 
41.7 \textcolor{gray}{(+\textbf{15.3})}\\
\checkmark &\checkmark &\checkmark  &
42.3 \textcolor{blue}{(+\textbf{10.1})}  & 
61.2 \textcolor{gray}{(+\textbf{13.7})}  & 
45.7 \textcolor{gray}{(+\textbf{11.3})} & 
26.7 \textcolor{gray}{(+\textbf{8.5})} & 
44.6 \textcolor{gray}{(+\textbf{9.4})} & 
57.6 \textcolor{gray}{(+\textbf{15.9})} \\
\bottomrule
\end{tabular}}
\end{center}
\vspace{-3mm}
\caption{Ablation studies on each components in Sparse R-CNN. Starting from Faster R-CNN, we gradually add learnable proposal boxes, iterative architecture, and dynamic head in Sparse R-CNN. All models are trained with set prediction loss.
}
\label{table5_2}
\vspace{-5mm}
\end{table*}

\myparagraph{Training details.}
ResNet-50~\cite{ResNet} is used as the backbone network unless otherwise specified. 
The optimizer is AdamW~\cite{loshchilov2018decoupled}  with weight decay 0.0001. 
The mini-batch is 16 images and all models are trained with 8 GPUs.
Default training schedule is 36 epochs and the initial learning rate is set to $2.5$ $\times$ $10^{-5}$, divided by 10 at epoch 27 and 33, respectively. 
The backbone is initialized with the pre-trained weights on ImageNet~\cite{ImageNet} and other newly added layers are initialized with Xavier~\cite{glorot2010understanding}. 
Data augmentation includes random horizontal,
scale jitter of resizing the input images such that the shortest side is at least 480 and at most 800 pixels while the longest at most 1333. Following~\cite{DETR, deformabledetr}, $\lambda_{cls}=2$, $\lambda_{L1}=5$, $\lambda_{giou}=2$. The default number of proposal boxes, proposal features and iteration is 100, 100 and 6, respectively. To stabilize training, the gradients are blocked at proposal boxes in each stage of iterative architecture, except initial proposal boxes.

\myparagraph{Inference details.}
The inference process is quite simple in \Ours. 
Given an input image, 
\Ours directly predicts 100 bounding boxes associated with their scores. 
The scores indicate the probability of boxes containing an object. 
For evaluation, we directly use these 100 boxes without any post-processing.

\subsection{Main Result}
We provide two versions of Sparse R-CNN for fair comparison with different detectors in Table~\ref{table5_1}.
The first one adopts 100 learnable proposal boxes \textit{without} random crop data augmentation,
and is used to make comparison with mainstream object detectors, \eg Faster R-CNN and RetinaNet~\cite{detectron2}. 
The second one leverages 300 learnable proposal boxes with random crop data augmentations, 
and is used to make comparison with DETR-series models~\cite{DETR,deformabledetr}.

As shown in Table~\ref{table5_1}, 
Sparse R-CNN outperforms well-established mainstream detectors, such as RetinaNet and Faster R-CNN, by a large margin. 
Surprisingly, Sparse R-CNN based on ResNet-50 achieves 42.8 AP, which has already competed with Faster R-CNN on ResNet-101 in accuracy. 
 
We note that DETR and Deformable DETR usually employ stronger feature extracting method, such as stacked encoder layers and deformable convolution. 
The stronger implementation of Sparse R-CNN is used to give a more fair comparison with these detectors. 
Sparse R-CNN exhibits higher accuracy 
even using the simple FPN as feature extracting method. 
Moreover, \Ours gets much better detection performance on small objects compared with DETR (26.7 AP vs. 22.5 AP).

\begin{table}[t]	
	\centering
    \setlength{\tabcolsep}{1.8mm}
\begin{tabular}{c c l c c}
\toprule
Cascade & Feature reuse & \multicolumn{1}{c}{AP} & AP$_{50}$ & AP$_{75}$ \\
\midrule
 & &18.5 & 35.0 & 17.7 
\\
\checkmark & &
20.5\textcolor{blue}{(+\textbf{2.0})} & 29.3 & 20.7
\\
\checkmark & \checkmark& 32.2\textcolor{blue}{(+\textbf{11.7})} & 47.5 & 34.4
\\
\bottomrule
\end{tabular}
	\caption{The effect of feature reuse in iterative architecture. Original cascading implementation makes no big difference. Concatenating object feature of previous stage to object feature of current stage leads to a huge improvement.}
    \label{table:cascade}
    \vspace{-5mm}
\end{table}

\begin{table}[t]	
	\centering
    \setlength{\tabcolsep}{1.8mm}
\begin{tabular}{c c l c c c c c }
\toprule
Self-att. &Ins. interact & \multicolumn{1}{c}{AP} & AP$_{50}$ & AP$_{75}$ \\
\midrule
&& 32.2 & 47.5 & 34.4\\
\checkmark& &37.2\textcolor{blue}{(+\textbf{5.0})} & 54.8 & 40.1\\
\checkmark&\checkmark & 42.3\textcolor{blue}{(+\textbf{5.1})} & 61.2 & 45.7\\
\bottomrule
\end{tabular}
	\caption{The effect of instance-interaction in dynamic head. Without instance interaction, dynamic head degenerates to self-attention. The gain comes from both self-attention and instance-interaction.}
    \label{table:selfatt}
    \vspace{-6mm}
\end{table}

The training convergence speed of Sparse R-CNN is $10\times$ faster over DETR, as shown in Figure~\ref{fig:curve}.
Since proposed, DETR has been suffering from slow convergence, which motivates the proposal of Deformable DETR.
Compared with Deformable DETR, Sparse R-CNN exhibits better performance in accuracy (45.0 AP vs. 43.8 AP) and shorter running-time (22 FPS vs. 19 FPS), with shorter training schedule (36 epochs vs. 50 epochs).

The inference time of Sparse R-CNN is on par with other detectors. 
We notice that the model with 100 proposals is running at 23 FPS, while 300 proposals only decreases to 22 FPS, thanks to the light design of the dynamic instance interactive head.

Table~\ref{table_test} compares Sparse R-CNN with other methods in COCO \texttt{test-dev} set. Using ResNeXt-101~\cite{resnext} as backbone, Sparse R-CNN achieves 46.9 AP without bells and whistles, 48.9 AP with DCN~\cite{DCN}. With additional test-time augmentations, Sparse R-CNN achieves 51.5 AP, on par with state-of-the-art methods.

\subsection{Module Analysis}
In this section, we analyze each component in Sparse R-CNN. 
All models are based on ResNet50-FPN backbone, 100 proposals, 3x training schedule, unless otherwise noted.

\myparagraph{Learnable proposal box.}
Starting with Faster R-CNN, we naively replace RPN with a sparse set of learnable proposal boxes. 
The performance drops from 40.2 AP (Table~\ref{table5_1} line 3) to 18.5 (Table~\ref{table5_2}).
We find that there is no noticeable improvement even more fully-connected layers are stacked.

\myparagraph{Iterative architecture.} 
Iteratively updating the boxes is an intuitive idea to improve its performance. 
However, we find that a simple cascade architecture does not make a big difference, as shown in Table~\ref{table:cascade}.
We analyze the reason is that compared with refined proposal boxes in ~\cite{CascadeRCNN} which mainly locating around the objects, the candidates in our case are much more coarse, making it hard to be optimized. We observe that the target object for one proposal box is usually consistent in the whole iterative process. 
Therefore, the object feature in previous stage can be reused to play a strong cue for the next stage, for example, the object feature encodes rich information such as object pose and location. To this end, we concatenate object feature of the previous stage to the current stage. 
This minor change of feature reuse results in a huge gain of 11.7 AP on basis of original cascade architecture. 
Finally, the iterative architecture brings 13.7 AP improvement, as shown in second row of Table~\ref{table5_2}.

\myparagraph{Dynamic head.}
The dynamic head uses object feature of previous stage in a different way with iterative architecture discussed above. 
Instead of simply concatenating, the object feature of previous stage is first processed by self-attention module, 
and then used as proposal feature to implement instance interaction of current stage. 
The self-attention module is applied to the set of object features for reasoning about the relation between objects. 
Table~\ref{table:selfatt} shows the benefit of self-attention and dynamic instance interaction. Finally, Sparse R-CNN achieves accuracy performance of 42.3 AP. 

\begin{table}[t]	
	\centering
    \centering
\begin{tabular}{c c c c c c c}
\toprule
Init. & AP & AP$_{50}$ & AP$_{75}$ & AP$_s$ & AP$_m$ & AP$_l$ \\
\midrule
Center & 41.5 &59.6 & 45.0 & 25.6 & 43.9 & 56.1 \\
Image & 42.3 & 61.2 & 45.7 & 26.7 & 44.6 & 57.6\\
Grid &  41.0 & 59.4 & 44.2 & 23.8 & 43.7 & 55.6\\
Random& 42.1 & 60.3 & 45.3 & 24.5 & 44.6 & 57.9 \\
\bottomrule
\end{tabular}

	\caption{Effect of initialization of proposal boxes. Detection performance is relatively robust to initialization of proposal boxes. }
    \label{table:boxinit}
    \vspace{-6mm}
\end{table}

\begin{table}[t]	
	\centering
	\small
    
\setlength{\tabcolsep}{2mm}
\begin{tabular}{c c c c c c}
\toprule
Proposals & AP & AP$_{50}$ & AP$_{75}$ & FPS & Training time \\
\midrule
100 & 42.3 & 61.2 & 45.7 & 23 & 19h \\
300 & 43.9 & 62.3 & 47.4 & 22 & 24h \\
500 & 44.6 & 63.2 & 48.5 & 20 & 60h \\
\bottomrule
\end{tabular}
	\caption{Effect of number of proposals. Increasing number of proposals leads to continuous improvement, while more proposals take more training time.}
    \label{table:proposal}
    \vspace{-6mm}
\end{table}

\begin{table}[t]	
	\centering
    \setlength{\tabcolsep}{2mm}
\begin{tabular}{c c c c c c}
\toprule
Stages & AP & AP$_{50}$ & AP$_{75}$ & FPS & Training time \\
\midrule
1 & 21.7  & 36.7 & 22.3 & 35 & 12h \\
2 & 36.2  & 52.8 & 38.8 & 33 & 13h \\
3 & 39.9  & 56.8 & 43.2 & 29 & 15h \\
6 & 42.3  & 61.2 & 45.7 & 23 & 19h \\
12 &41.6  & 60.2 & 45.0 & 17 & 30h \\
\bottomrule
\end{tabular}
	\caption{Effect of number of stages. Gradually increasing the number of stages, the performance is saturated at 6 stages.}
    \label{table:stage}
    \vspace{-7mm}
\end{table}

\myparagraph{Initialization of proposal boxes.}
The dense detectors always heavily depend on design of object candidates, whereas, object candidates in Sparse R-CNN are learnable and thus, all efforts related to designing hand-crafted anchors are avoided. 
However, one may concern that the initialization of proposal boxes plays a key role in \Ours. Here we study the effect of different methods for initializing proposal boxes: 
\begin{itemize}
    \item ``Center" means all proposal boxes are located in the center of image at beginning, height and width is set to 0.1 of image size.
    \item ``Image" means all proposal boxes are initialized as the whole image size. 
    \item ``Grid" means proposal boxes are initialized as regular grid in image, which is exactly the initial boxes in G-CNN~\cite{gcnn}. 
    \item ``Random" denotes the center, height and width of proposal boxes are randomly initialized with Gaussian distribution. 
\end{itemize}
From Table~\ref{table:boxinit} we show that the final performance of \Ours is relatively robust to the initialization of proposal boxes. 

\myparagraph{Number of proposals.}
The number of proposals largely effects both dense and sparse detectors. 
Original Faster R-CNN uses 300 proposals~\cite{FasterRCNN}. 
Later on it increases to 2000~\cite{detectron2} and obtains better performance. 
We also study the effect of proposal numbers on Sparse R-CNN in Table~\ref{table:proposal}. 
Increasing proposal number from 100 to 500 leads to continuous improvement, 
indicating that our framework is easily to be used in various circumstances. Whereas, 500 proposals take much more training time, so we choose 100 and 300 as the main configurations.

\myparagraph{Number of stages in iterative architecture.}
Iterative architecture is a widely-used technique to improve object detection performance~\cite{CascadeRCNN,DETR,boosted}, especially for Sparse R-CNN. Table~\ref{table:stage} shows the effect of stage numbers in iterative architecture. Without iterative architecture, performance is merely 21.7 AP. 
Considering the input proposals of first stage is a guess of possible object positions, this result is not surprising. 
Increasing to 2 stage brings in a gain of 14.5 AP, up to competitive 36.2 AP. 
Gradually increasing the number of stages, the performance is saturated at 6 stages. We choose 6 stages as the default configuration.

\begin{table}[t]	
	\centering
    \setlength{\tabcolsep}{1.5mm}
\begin{tabular}{l l c c}
\toprule
Method  & AP & AP$_{50}$ & AP$_{75}$ \\
\midrule
Multi-head Attention~\cite{vaswani2017attention} & 35.7 & 54.9 & 37.7\\
Dynamic head& 42.3\textcolor{blue}{(+\textbf{6.6})} & 61.2 & 45.7 \\
\bottomrule
\end{tabular}
	\caption{Dynamic head vs. Multi-head Attention. As object recognition head, dynamic head outperforms multi-head attention.}
    \label{table:mha}
    \vspace{-6mm}
\end{table}

\begin{table}[t]	
	\centering
    \setlength{\tabcolsep}{0.8mm}
\begin{tabular}{l c l c c}
\toprule
Method & Pos. encoding & AP & AP$_{50}$ & AP$_{75}$ \\
\midrule
DETR~\cite{DETR} & \checkmark & 40.6  & 61.6 & -\\
DETR~\cite{DETR} & & 32.8 \textcolor{blue}{(-\textbf{7.8})}  & 55.2 & - \\
Sparse R-CNN& \checkmark & 41.9 & 60.9 & 45.0 \\
Sparse R-CNN & & 42.3\textcolor{blue}{(+\textbf{0.4})} & 61.2 & 45.7 \\
\bottomrule
\end{tabular}
	\caption{Proposal feature vs. Object query. Object query is learned positional encoding, while proposal feature is irrelevant to position.}
    \label{table:oq}
    \vspace{-6mm}
\end{table}

\begin{figure*}[!t]
\vspace{5mm}
\includegraphics[width=0.99\textwidth]{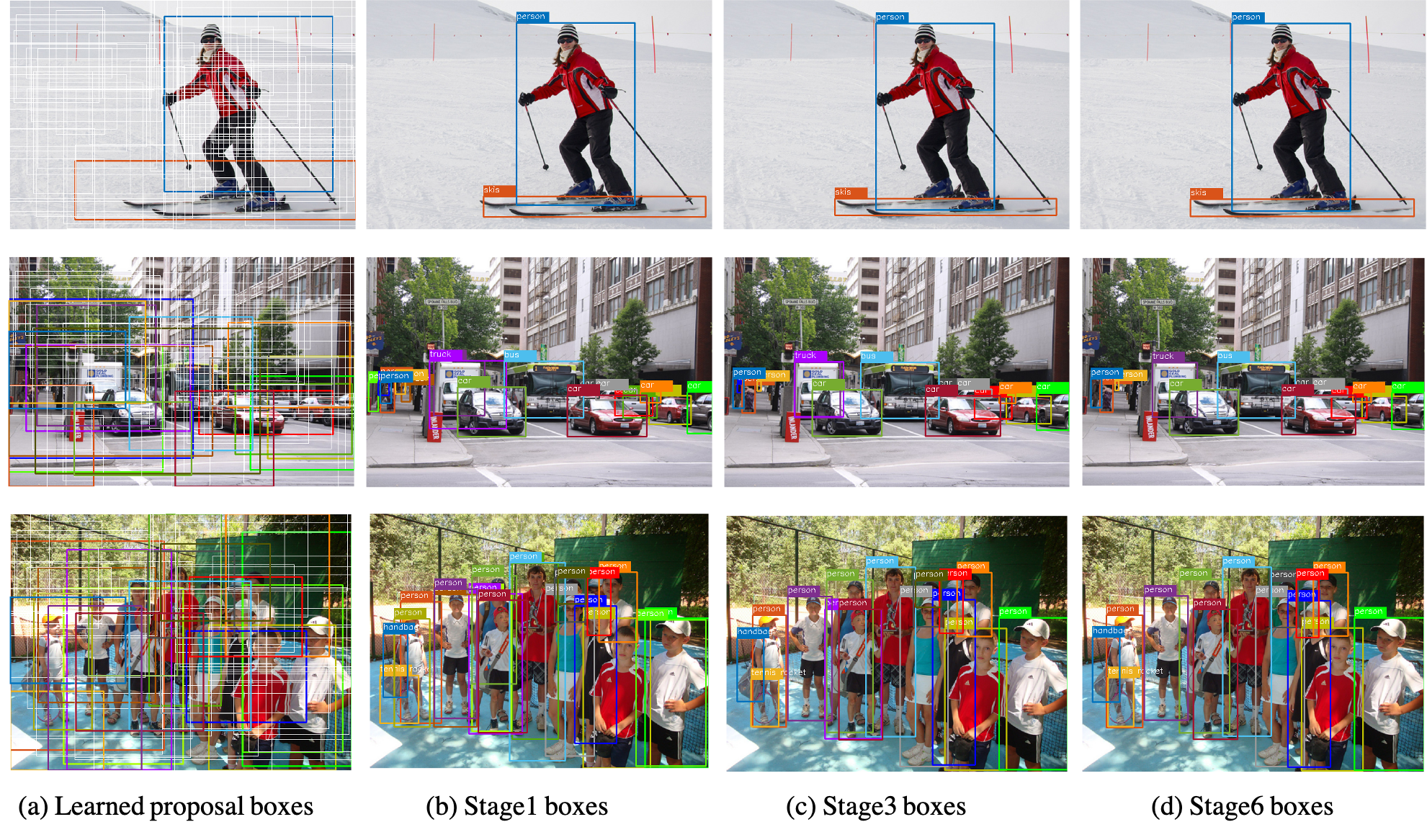}
\caption{Visualization of predicted boxes of each stage in iterative architecture, including learned proposal boxes. Learned proposal boxes are drawn in white color, except those are shown in later stages.
Predicted boxes of classification score above $0.3$ are shown. The boxes from the same proposal are drawn in the same color. 
The learned proposal boxes are randomly distributed on the image and together cover the whole image. 
The iterative heads gradually refine box position and remove duplicate ones.} 
\label{fig:box}
\end{figure*}

\myparagraph{Dynamic head vs. Multi-head Attention.} As discussed in Section~\ref{sec:3}, dynamic head uses proposal feature to filter RoI feature and finally outputs object feature.
We find that multi-head attention module~\cite{vaswani2017attention} provides another possible implementation for the instance interaction.
We carry out the comparison experiments in Table~\ref{table:mha}, and its performance falls behind 6.6 AP. 
Compared with linear multi-head attention, our dynamic head is much more flexible, whose parameters are conditioned on its specific proposal feature and more non-linear capacity can be easily introduced.

\myparagraph{Proposal feature vs. Object query.}
Object query proposed in DETR~\cite{DETR} shares a similar design as proposal feature. Here we make a comparison of object query~\cite{DETR} proposed in DETR and our proposal feature.
As discussed in~\cite{DETR}, object query is learned positional encoding, guiding the decoder interacting with the summation of image feature map and spatial positional encoding. 
Using only image feature map will lead to a significant drop. 
However, our proposal feature can be seen as a feature filter, 
which is irrelevant to position. 
The comparisons are shown in Table~\ref{table:oq}, DETR drops 7.8 AP if the spatial positional encoding is removed. 
On the contrary, positional encoding gives no gain in Sparse R-CNN.

\subsection{The Proposal Boxes Behavior} 

Figure~\ref{fig:box} shows the learned proposal boxes of a converged model. 
These boxes are randomly distributed on the image to cover the whole image area. 
This guarantees the recall performance on the condition of sparse candidates. 
Further, each stage of cascading heads gradually refines box position and remove duplicate ones. 
This results in high precision performance. 
Figure~\ref{fig:box} also shows that Sparse R-CNN presents robust performance in both rare and crowd scenarios. 
For object in rare scenario, its duplicate boxes are removed within a few of stages. 
Crowd scenarios consume more stages to refine but finally each object is detected precisely and uniquely.

\section{Conclusion}
We present Sparse R-CNN, a purely sparse method for object detection in images. A fixed sparse set of learned object proposals are provided to perform classification and location by dynamic heads. Final predictions are directly output without non-maximum suppression post-procedure. Sparse R-CNN demonstrates its accuracy, run-time and training  convergence performance on par with the well-established detector. We hope our work could inspire re-thinking the convention of dense prior and exploring next generation of object detector.

\vspace{5mm}
\noindent
\textbf{Acknowledgements} This work was supported by the General Research Fund of HK No.27208720.

\newpage
\appendix
\begin{center}{\bf \Large Appendix}\end{center}\vspace{-2mm}
\section{Crowded Scene}
One concern about Sparse R-CNN is its performance on crowded scene. We conduct experiments on CrowdHuman~\cite{crowdhuman}, a highly crowded human detection benchmark. Following~\cite{citypersons,detrhuman,cvpods}, we use evaluation metrics as AP, mMR and Recall under IoU 50. 

On CrowdHuman, Sparse R-CNN is trained for 50 epochs, with learning rate divided by 10 at epoch 40. The proposal number is 500. The shortest side of input image is at least 480 and at most 800, while longest side is at most 1500. Other details are the same as COCO dataset.

\begin{table}[H]
\begin{center}
\setlength{\tabcolsep}{1.5mm}
\begin{tabular}
{
l | c | c c c 
}
\toprule
Method & 
NMS &
AP  & 
mMR  & 
Recall \\
\midrule

Faster R-CNN~\cite{FasterRCNN} & \checkmark &  85.0 & 50.4 & 90.2 \\
RetinaNet~\cite{FocalLoss} & \checkmark & 81.7 & 57.6 & 88.6 \\
FCOS~\cite{FCOS} & \checkmark &  86.1 & 55.2 & 94.3 \\
DETR~\cite{DETR} & $\circ$ & 66.1 & 80.6 & - \\
Deformable DETR~\cite{deformabledetr} & $\circ$ & 86.7 & 54.0 & 92.5 \\

\midrule
Sparse R-CNN & $\circ$ &  \textbf{89.2} & \textbf{48.3} & \textbf{95.9} 
\\

\bottomrule
\end{tabular}
\end{center}
\vspace{-2mm}
\caption{Performance of different detectors on CrowdHuman dataset. All models are trained on \texttt{train} split ($\sim$15k images) and evaluated on \texttt{val} split ($\sim$4k images).}
\label{table:crowd}
\vspace{-4mm}
\end{table}

From Table~\ref{table:crowd}, we are surprised to see that Sparse R-CNN achieves better performance than well-established mainstream detectors, such as Faster R-CNN, RetinaNet and FCOS. Meanwhile, Sparse R-CNN improves 23.1 and 2.5 AP than DETR and Deformable DETR, two recent end-to-end detectors. 

The experiments on CrowdHuman show that Sparse R-CNN is also applicable on crowded scene. We hope Sparse R-CNN could serve as a solid baseline used in various detection scenarios. 

\section{Self-supervised Pre-training}
Object detection has adopted ImageNet-supervised pre-training weight~\cite{ImageNet,RCNN} for past several years. Recently, self-supervised methods show promising benefits than the supervised counterparts on well-established detectors~\cite{simclr,moco,byol,swav}. Accordingly, we list experiment results of self-supervised methods on Sparse R-CNN, such as DetCo~\cite{detco}, SCRL~\cite{scrl}, where DetCo introduces contrastive learning between global image and local patches, SCRL learns spatially consistent representations of randomly cropped local regions by geometric translations and zooming operations.

In Table~\ref{table:weight}, Sparse R-CNN obtains consistent improvement by replacing ImageNet-supervised pre-training weight to self-supervised ones. 

\begin{table}[t]
\begin{center}
{\setlength{\tabcolsep}{0.8mm}
\begin{tabular}
{
l l c c c c c c c c 
}
\toprule
Method & 
\multicolumn{1}{c}{AP} & 
\multicolumn{1}{c}{AP$_{50}$}  & 
\multicolumn{1}{c}{AP$_{75}$}  & 
\multicolumn{1}{c}{AP$_s$}  & 
\multicolumn{1}{c}{AP$_m$}  & 
\multicolumn{1}{c}{AP$_l$}  & \\
\midrule

Supervised~\cite{ImageNet} & 
45.0 & 63.4 & 48.2 &26.9 & 47.2 & 59.5 \\

\midrule

DetCo~\cite{detco} & 
46.5\textcolor{blue}{(+\textbf{1.5})} &  65.7 & 50.8 & 30.8 & 49.5 & 59.7\\

SCRL~\cite{scrl} & 
46.7\textcolor{blue}{(+\textbf{1.7})} & 65.7 & 51.1 & - & - & - \\

\bottomrule
\end{tabular}}
\end{center}
\vspace{-3mm}
\caption{Comparisons of supervised and self-supervised pre-training weights on Sparse R-CNN. All models use ResNet-50 as backbone.}
\label{table:weight}
\vspace{-4mm}
\end{table}

\section{Backbone Architecture}

The default backbone of Sparse R-CNN is ResNet-50, CNN-based architecture. Recently, Transformer-based architecture achieves great success in computer vision community~\cite{vaswani2017attention,wuvit,vit}. We list Sparse R-CNN performance with two recently-proposed Transformer backbone, PVT~\cite{pvt} and Swin Transformer~\cite{swin}, where PVT applies a progressive shrinking pyramid structure, Swin Transformer constructs hierarchical representation computed with shifted window.

From Table~\ref{table:arc}, both Transformer-based backbones achieve better performance than CNN on Sparse R-CNN.

\begin{table}[H]
\begin{center}
{\setlength{\tabcolsep}{1.2mm}
\begin{tabular}
{
l l c c c c c c c c 
}
\toprule
Method & 
\multicolumn{1}{c}{AP} & 
\multicolumn{1}{c}{AP$_{50}$}  & 
\multicolumn{1}{c}{AP$_{75}$}  & 
\multicolumn{1}{c}{AP$_s$}  & 
\multicolumn{1}{c}{AP$_m$}  & 
\multicolumn{1}{c}{AP$_l$}  & \\
\midrule

CNN~\cite{ResNet} & 
45.0 & 63.4 & 48.2 &26.9 & 47.2 & 59.5 \\

\midrule

PVT~\cite{pvt} & 
45.7\textcolor{blue}{(+\textbf{0.7})} & - & - & -& - & -  \\

Swin~\cite{swin} & 
47.9\textcolor{blue}{(+\textbf{2.9})} & 67.3 & 52.3 & - & - & - \\
  
\bottomrule
\end{tabular}}
\end{center}
\vspace{-3mm}
\caption{Comparisons of CNN and Transformer backbone on Sparse R-CNN.}
\label{table:arc}
\vspace{-4mm}
\end{table}

{\small
\bibliographystyle{ieee_fullname}
\bibliography{egbib}

\begin{thebibliography}{10}\itemsep=-1pt

\bibitem{Soft-NMS}
Navaneeth Bodla, Bharat Singh, Rama Chellappa, and Larry~S. Davis.
\newblock {Soft-NMS} -- improving object detection with one line of code.
\newblock In {\em ICCV}, 2017.

\bibitem{CascadeRCNN}
Zhaowei Cai and Nuno Vasconcelos.
\newblock Cascade {R-CNN}: Delving into high quality object detection.
\newblock In {\em CVPR}, 2018.

\bibitem{DETR}
Nicolas Carion, Francisco Massa, Gabriel Synnaeve, Nicolas Usunier, Alexander
  Kirillov, and Sergey Zagoruyko.
\newblock {End-to-End} object detection with transformers.
\newblock In {\em ECCV}, 2020.

\bibitem{swav}
Mathilde Caron, Ishan Misra, Julien Mairal, Priya Goyal, Piotr Bojanowski, and
  Armand Joulin.
\newblock Unsupervised learning of visual features by contrasting cluster
  assignments.
\newblock {\em arXiv preprint arXiv:2006.09882}, 2020.

\bibitem{simclr}
Ting Chen, Simon Kornblith, Mohammad Norouzi, and Geoffrey Hinton.
\newblock A simple framework for contrastive learning of visual
  representations.
\newblock {\em arXiv preprint arXiv:2002.05709}, 2020.

\bibitem{R-FCN}
Jifeng Dai, Yi Li, Kaiming He, and Jian Sun.
\newblock {R-FCN}: Object detection via region-based fully convolutional
  networks.
\newblock In {\em NeurIPS}, 2016.

\bibitem{DCN}
Jifeng Dai, Haozhi Qi, Yuwen Xiong, Yi Li, Guodong Zhang, Han Hu, and Yichen
  Wei.
\newblock Deformable convolutional networks.
\newblock In {\em ICCV}, 2017.

\bibitem{hog}
Navneet Dalal and Bill Triggs.
\newblock Histograms of oriented gradients for human detection.
\newblock In {\em CVPR}, 2005.

\bibitem{ImageNet}
Jia Deng, Wei Dong, Richard Socher, Li-Jia Li, Kai Li, and Li Fei-Fei.
\newblock {ImageNet}: A large-scale hierarchical image database.
\newblock In {\em CVPR}, 2009.

\bibitem{vit}
Alexey Dosovitskiy, Lucas Beyer, Alexander Kolesnikov, Dirk Weissenborn,
  Xiaohua Zhai, Thomas Unterthiner, Mostafa Dehghani, Matthias Minderer, Georg
  Heigold, Sylvain Gelly, et~al.
\newblock An image is worth 16x16 words: Transformers for image recognition at
  scale.
\newblock {\em arXiv preprint arXiv:2010.11929}, 2020.

\bibitem{PASCAL-VOC}
Mark Everingham, Luc. Van~Gool, Christopher K.~I. Williams, John Winn, and
  Andrew Zisserman.
\newblock The pascal visual object classes ({VOC}) challenge.
\newblock {\em IJCV}, 88(2):303--338, 2010.

\bibitem{dpm}
Pedro Felzenszwalb, Ross Girshick, David McAllester, and Deva Ramanan.
\newblock Object detection with discriminatively trained part based models.
\newblock {\em T-PAMI}, 32(9):1627--1645, 2010.

\bibitem{FastRCNN}
Ross Girshick.
\newblock Fast {R-CNN}.
\newblock In {\em ICCV}, 2015.

\bibitem{RCNN}
Ross Girshick, Jeff Donahue, Trevor Darrell, and Jitendra Malik.
\newblock Rich feature hierarchies for accurate object detection and semantic
  segmentation.
\newblock In {\em CVPR}, 2014.

\bibitem{glorot2010understanding}
Xavier Glorot and Yoshua Bengio.
\newblock Understanding the difficulty of training deep feedforward neural
  networks.
\newblock In {\em Proceedings of the thirteenth international conference on
  artificial intelligence and statistics}, pages 249--256, 2010.

\bibitem{byol}
Jean-Bastien Grill, Florian Strub, Florent Altch{\'e}, Corentin Tallec,
  Pierre~H Richemond, Elena Buchatskaya, Carl Doersch, Bernardo~Avila Pires,
  Zhaohan~Daniel Guo, Mohammad~Gheshlaghi Azar, et~al.
\newblock Bootstrap your own latent: A new approach to self-supervised
  learning.
\newblock {\em arXiv preprint arXiv:2006.07733}, 2020.

\bibitem{moco}
Kaiming He, Haoqi Fan, Yuxin Wu, Saining Xie, and Ross Girshick.
\newblock Momentum contrast for unsupervised visual representation learning.
\newblock In {\em Proceedings of the IEEE/CVF Conference on Computer Vision and
  Pattern Recognition}, pages 9729--9738, 2020.

\bibitem{MaskRCNN}
Kaiming He, Georgia Gkioxari, Piotr Dollar, and Ross Girshick.
\newblock Mask {R-CNN}.
\newblock In {\em ICCV}, 2017.

\bibitem{ResNet}
Kaiming He, Xiangyu Zhang, Shaoqing Ren, and Jian Sun.
\newblock Deep residual learning for image recognition.
\newblock In {\em CVPR}, 2016.

\bibitem{RelationNetworks}
Han Hu, Jiayuan Gu, Zheng Zhang, Jifeng Dai, and Yichen Wei.
\newblock Relation networks for object detection.
\newblock In {\em CVPR}, 2018.

\bibitem{DenseBox}
Lichao Huang, Yi Yang, Yafeng Deng, and Yinan Yu.
\newblock {DenseBox}: Unifying landmark localization with end to end object
  detection.
\newblock {\em arXiv preprint arXiv:1509.04874}, 2015.

\bibitem{BatchNorm}
Sergey Ioffe and Christian Szegedy.
\newblock Batch normalization: Accelerating deep network training by reducing
  internal covariate shift.
\newblock In {\em ICML}, 2015.

\bibitem{jia2016dynamic}
Xu Jia, Bert De~Brabandere, Tinne Tuytelaars, and Luc~V Gool.
\newblock Dynamic filter networks.
\newblock In {\em NIPS}, pages 667--675, 2016.

\bibitem{foveabox}
Tao Kong, Fuchun Sun, Huaping Liu, Yuning Jiang, Lei Li, and Jianbo Shi.
\newblock Foveabox: Beyound anchor-based object detection.
\newblock {\em IEEE Transactions on Image Processing}, 29:7389--7398, 2020.

\bibitem{AlexNet}
Alex Krizhevsky, Ilya Sutskever, and Geoffrey~E Hinton.
\newblock {ImageNet} classification with deep convolutional neural networks.
\newblock In {\em NeurIPS}, 2012.

\bibitem{CornerNet}
Hei Law and Jia Deng.
\newblock {CornerNet}: Detecting objects as paired keypoints.
\newblock In {\em ECCV}, 2018.

\bibitem{detrhuman}
Matthieu Lin, Chuming Li, Xingyuan Bu, Ming Sun, Chen Lin, Junjie Yan, Wanli
  Ouyang, and Zhidong Deng.
\newblock Detr for pedestrian detection.
\newblock {\em arXiv preprint arXiv:2012.06785}, 2020.

\bibitem{FPN}
Tsung-Yi Lin, Piotr Dollar, Ross Girshick, Kaiming He, Bharath Hariharan, and
  Serge Belongie.
\newblock Feature pyramid networks for object detection.
\newblock In {\em CVPR}, 2017.

\bibitem{FocalLoss}
Tsung-Yi Lin, Priya Goyal, Ross Girshick, Kaiming He, and Piotr Dollar.
\newblock Focal loss for dense object detection.
\newblock In {\em ICCV}, 2017.

\bibitem{COCO}
Tsung-Yi Lin, Michael Maire, Serge Belongie, James Hays, Pietro Perona, Deva
  Ramanan, Piotr Dollár, and C.~Lawrence Zitnick.
\newblock Microsoft {COCO}: Common objects in context.
\newblock In {\em ECCV}, 2014.

\bibitem{SSD}
Wei Liu, Dragomir Anguelov, Dumitru Erhan, Christian Szegedy, Scott Reed,
  Cheng-Yang Fu, and Alexander~C. Berg.
\newblock {SSD}: Single shot multibox detector.
\newblock In {\em ECCV}, 2016.

\bibitem{swin}
Ze Liu, Yutong Lin, Yue Cao, Han Hu, Yixuan Wei, Zheng Zhang, Stephen Lin, and
  Baining Guo.
\newblock Swin transformer: Hierarchical vision transformer using shifted
  windows.
\newblock {\em arXiv preprint arXiv:2103.14030}, 2021.

\bibitem{loshchilov2018decoupled}
Ilya Loshchilov and Frank Hutter.
\newblock Decoupled weight decay regularization.
\newblock In {\em International Conference on Learning Representations}, 2018.

\bibitem{gcnn}
Mahyar Najibi, Mohammad Rastegari, and Larry~S Davis.
\newblock G-cnn: an iterative grid based object detector.
\newblock In {\em Proceedings of the IEEE conference on computer vision and
  pattern recognition}, pages 2369--2377, 2016.

\bibitem{YOLO}
Joseph Redmon, Santosh Divvala, Ross Girshick, and Ali Farhadi.
\newblock You only look once: Unified, real-time object detection.
\newblock In {\em CVPR}, 2016.

\bibitem{YOLO9000}
Joseph Redmon and Ali Farhadi.
\newblock {YOLO9000}: Better, faster, stronger.
\newblock In {\em CVPR}, 2017.

\bibitem{FasterRCNN}
Shaoqing Ren, Kaiming He, Ross Girshick, and Jian Sun.
\newblock Faster {R-CNN}: Towards real-time object detection with region
  proposal networks.
\newblock In {\em NeurIPS}, 2015.

\bibitem{GIoU}
Hamid Rezatofighi, Nathan Tsoi, JunYoung Gwak, Amir Sadeghian, Ian Reid, and
  Silvio Savarese.
\newblock Generalized intersection over union: A metric and a loss for bounding
  box regression.
\newblock In {\em CVPR}, 2019.

\bibitem{scrl}
Byungseok Roh, Wuhyun Shin, Ildoo Kim, and Sungwoong Kim.
\newblock Spatially consistent representation learning.
\newblock {\em arXiv preprint arXiv:2103.06122}, 2021.

\bibitem{OverFeat}
Pierre Sermanet, David Eigen, Xiang Zhang, Michael Mathieu, Robert Fergus, and
  Yann Lecun.
\newblock {OverFeat}: Integrated recognition, localization and detection using
  convolutional networks.
\newblock In {\em ICLR}, 2014.

\bibitem{crowdhuman}
Shuai Shao, Zijian Zhao, Boxun Li, Tete Xiao, Gang Yu, Xiangyu Zhang, and Jian
  Sun.
\newblock Crowdhuman: A benchmark for detecting human in a crowd.
\newblock {\em arXiv preprint arXiv:1805.00123}, 2018.

\bibitem{stewart2016end}
Russell Stewart, Mykhaylo Andriluka, and Andrew~Y Ng.
\newblock End-to-end people detection in crowded scenes.
\newblock In {\em Proceedings of the IEEE conference on computer vision and
  pattern recognition}, pages 2325--2333, 2016.

\bibitem{EfficientDet}
Mingxing Tan, Ruoming Pang, and Quoc~V. Le.
\newblock {EfficientDet}: Scalable and efficient object detection.
\newblock In {\em CVPR}, 2020.

\bibitem{condinst}
Zhi Tian, Chunhua Shen, and Hao Chen.
\newblock Conditional convolutions for instance segmentation.
\newblock {\em arXiv preprint arXiv:2003.05664}, 2020.

\bibitem{FCOS}
Zhi Tian, Chunhua Shen, Hao Chen, and Tong He.
\newblock {FCOS}: Fully convolutional one-stage object detection.
\newblock In {\em ICCV}, 2019.

\bibitem{SelectiveSearch}
Jasper~RR Uijlings, Koen~EA Van De~Sande, Theo Gevers, and Arnold~WM Smeulders.
\newblock Selective search for object recognition.
\newblock {\em IJCV}, 104(2):154--171, 2013.

\bibitem{vaswani2017attention}
Ashish Vaswani, Noam Shazeer, Niki Parmar, Jakob Uszkoreit, Llion Jones,
  Aidan~N Gomez, {\L}ukasz Kaiser, and Illia Polosukhin.
\newblock Attention is all you need.
\newblock In {\em Advances in neural information processing systems}, pages
  5998--6008, 2017.

\bibitem{boosted}
Paul Viola and Michael Jones.
\newblock Rapid object detection using a boosted cascade of simple features.
\newblock In {\em Proceedings of the 2001 IEEE computer society conference on
  computer vision and pattern recognition. CVPR 2001}, volume~1, pages I--I.
  IEEE, 2001.

\bibitem{yolov4scale}
Chien-Yao Wang, Alexey Bochkovskiy, and Hong-Yuan~Mark Liao.
\newblock Scaled-yolov4: Scaling cross stage partial network.
\newblock {\em arXiv preprint arXiv:2011.08036}, 2020.

\bibitem{pvt}
Wenhai Wang, Enze Xie, Xiang Li, Deng-Ping Fan, Kaitao Song, Ding Liang, Tong
  Lu, Ping Luo, and Ling Shao.
\newblock Pyramid vision transformer: A versatile backbone for dense prediction
  without convolutions.
\newblock {\em arXiv preprint arXiv:2102.12122}, 2021.

\bibitem{SOLOv2}
Xinlong Wang, Rufeng Zhang, Tao Kong, Lei Li, and Chunhua Shen.
\newblock Solov2: Dynamic and fast instance segmentation.
\newblock In {\em NIPS}, 2020.

\bibitem{wuvit}
Bichen Wu, Chenfeng Xu, Xiaoliang Dai, Alvin Wan, Peizhao Zhang, Zhicheng Yan,
  Masayoshi Tomizuka, Joseph Gonzalez, Kurt Keutzer, and Peter Vajda.
\newblock Visual transformers: Token-based image representation and processing
  for computer vision.
\newblock {\em arXiv preprint arXiv:2006.03677}, 2020.

\bibitem{detectron2}
Yuxin Wu, Alexander Kirillov, Francisco Massa, Wan-Yen Lo, and Ross Girshick.
\newblock Detectron2.
\newblock \url{https://github.com/facebookresearch/detectron2}, 2019.

\bibitem{detco}
Enze Xie, Jian Ding, Wenhai Wang, Xiaohang Zhan, Hang Xu, Zhenguo Li, and Ping
  Luo.
\newblock Detco: Unsupervised contrastive learning for object detection.
\newblock {\em arXiv preprint arXiv:2102.04803}, 2021.

\bibitem{resnext}
Saining Xie, Ross Girshick, Piotr Doll{\'a}r, Zhuowen Tu, and Kaiming He.
\newblock Aggregated residual transformations for deep neural networks.
\newblock In {\em Proceedings of the IEEE conference on computer vision and
  pattern recognition}, pages 1492--1500, 2017.

\bibitem{yang2019learning3d}
Bo Yang, Jianan Wang, Ronald Clark, Qingyong Hu, Sen Wang, Andrew Markham, and
  Niki Trigoni.
\newblock Learning object bounding boxes for 3d instance segmentation on point
  clouds.
\newblock In {\em Advances in Neural Information Processing Systems}, pages
  6740--6749, 2019.

\bibitem{reppoints}
Ze Yang, Shaohui Liu, Han Hu, Liwei Wang, and Stephen Lin.
\newblock {RepPoints}: Point set representation for object detection.
\newblock In {\em ICCV}, 2019.

\bibitem{DynamicRCNN}
Hongkai Zhang, Hong Chang, Bingpeng Ma, Naiyan Wang, and Xilin Chen.
\newblock Dynamic {R-CNN}: Towards high quality object detection via dynamic
  training.
\newblock In {\em ECCV}, 2020.

\bibitem{citypersons}
Shanshan Zhang, Rodrigo Benenson, and Bernt Schiele.
\newblock Citypersons: A diverse dataset for pedestrian detection.
\newblock In {\em Proceedings of the IEEE Conference on Computer Vision and
  Pattern Recognition}, pages 3213--3221, 2017.

\bibitem{ATSS}
Shifeng Zhang, Cheng Chi, Yongqiang Yao, Zhen Lei, and Stan~Z. Li.
\newblock Bridging the gap between anchor-based and anchor-free detection via
  adaptive training sample selection.
\newblock In {\em CVPR}, 2020.

\bibitem{CenterNet}
Xingyi Zhou, Dequan Wang, and Philipp Kr{\"{a}}henb{\"{u}}hl.
\newblock Objects as points.
\newblock {\em arXiv preprint arXiv:1904.07850}, 2019.

\bibitem{cvpods}
Benjin Zhu*, Feng Wang*, Jianfeng Wang, Siwei Yang, Jianhu Chen, and Zeming Li.
\newblock cvpods: All-in-one toolbox for computer vision research, 2020.

\bibitem{deformabledetr}
Xizhou Zhu, Weijie Su, Lewei Lu, Bin Li, Xiaogang Wang, and Jifeng Dai.
\newblock Deformable detr: Deformable transformers for end-to-end object
  detection.
\newblock {\em arXiv preprint arXiv:2010.04159}, 2020.

\end{thebibliography}
}

\end{document}